\documentclass[acmlarge]{acmart}

\AtBeginDocument{%
  }

\newcommand{\draftonly}[1]{#1} 
\newcommand{\draftcomment}[3]{\draftonly{{\textcolor{#3}{[\textbf{#1--\textsc{#2}}]}}}}

\newcommand{\rucomment}[1]{\draftcomment{#1}{Rheeya}{magenta}}

\usepackage{caption}
\usepackage{subcaption}
\usepackage{graphicx}
\graphicspath{{./Images/}}

\begin{document}

\title{Improving Bilingual Capabilities of Language Models to Support Diverse Linguistic Practices in Education}

\author{Anand Syamkumar}
\affiliation{%
\institution{Stony Brook University}
\city{Stony Brook}
\state{New York}
\country{USA}
}
\authornote{Both authors contributed equally to this research.}
\author{Nora Tseng}
\authornotemark[1]
\affiliation{%
\institution{University of Wisconsin-Madison}
\city{Madison}
\state{Wisconsin}
\country{USA}
}
\author{Kaycie Barron}
\affiliation{%
\institution{University of Wisconsin-Madison}
\city{Madison}
\state{Wisconsin}
\country{USA}
}
\author{Shanglin Yang}
\affiliation{%
\institution{University of Wisconsin-Madison}
\city{Madison}
\state{Wisconsin}
\country{USA}
}
\author{Shamya Karumbaiah}
\affiliation{%
\institution{University of Wisconsin-Madison}
\city{Madison}
\state{Wisconsin}
\country{USA}
}
\author{Rheeya Uppal}
\affiliation{%
\institution{University of Wisconsin-Madison}
\city{Madison}
\state{Wisconsin}
\country{USA}
}
\author{Junjie Hu}
\affiliation{%
\institution{University of Wisconsin-Madison}
\city{Madison}
\state{Wisconsin}
\country{USA}
}
\renewcommand{\shortauthors}{Syamkumar, Tseng et al.}

\begin{abstract}

Large language models (LLMs) offer promise in generating educational content, providing instructor feedback, and reducing teacher workload on assessments. While prior studies have focused on studying LLM-powered learning analytics, limited research has examined how effective LLMs are in a bilingual context. In this paper, we study the effectiveness of multilingual large language models (MLLMs) across monolingual (English-only, Spanish-only) and bilingual (Spanglish) student writing. We present a learning analytics use case that details LLM performance in assessing acceptable and unacceptable explanations of Science and Social Science concepts. Our findings reveal a significant bias in the grading performance of pre-trained models for bilingual writing compared to English-only and Spanish-only writing. Following this, we fine-tune open-source MLLMs including Llama 3.1 and Mistral NeMo using synthetic datasets generated in English, Spanish, and Spanglish. Our experiments indicate that the models perform significantly better for all three languages after fine-tuning with bilingual data. This study highlights the potential of enhancing MLLM effectiveness to support authentic language practices amongst bilingual learners. It also aims to illustrate the value of incorporating non-English languages into the design and implementation of language models in education.

\end{abstract}




\maketitle
\section{Introduction}
\label{sec:intro}

Recent research in learning analytics highlights the importance of addressing diversity, inclusivity, and equity to better support all learners~\cite{karumbaiah2021using, shams2023ai}. The advent of large language models (i.e., algorithms trained on large amounts of text to understand and process human language) in learning analytics research and practice raises similar questions about equity~\cite{anis2023leveraging}. How well can language models trained predominantly on mainstream English data serve students from diverse linguistic backgrounds?

Multilingual large language models (MLLMs) that can process and produce text in multiple languages offer promising new directions for supporting authentic multilingual communication. However, despite their impressive performance in individual languages, MLLMs are limited in their ability to switch fluidly across languages (known as translanguaging or code-switching~\cite{bang2023multitask, zhang2023multilingual}) - leading to an inaccurate representation of language use by bilingual learners. Homogeneity in MLLM training~\cite{yong2023prompting} due to scarcities in code-switched data further contributes to the issue, as well as the lack of safety benchmarks and comprehensive evaluation~\cite{qin2024multilingual}. As learning analytics powered by language models make their way into classrooms, studying the affordances and constraints of using MLLMs becomes increasingly important.Building upon existing research on translanguaging, bi/multilingualism, and natural language processing, the current study necessitates the integration of languages produced by bilingual learners, such as Spanglish (translanguaging in Spanish and English) into the training and evaluation of MLLMs. We investigate the code-switching and multilingual capabilities of MLLMs in an illustrative learning analytics use case of assessing student writing. We ask the following research question: \textit{How does MLLM performance vary across assessing monolingual (English, Spanish) and bilingual (Spanglish) student writing?}

We first evaluate the performance of two MLLMs in assessing Science and Social Science ideas as expressed in English, Spanish, and Spanglish. To overcome the data scarcity issue, we create and use a synthetically-generated dataset evaluated by humans on both the language and content accuracy. Then, we attempt to improve MLLM performance using techniques such as prompting and fine-tuning. Our hypotheses include:
\begin{itemize}
    \item H1: MLLMs are significantly more accurate when ideas are presented in English or Spanish, but are significantly less accurate when ideas are in Spanglish.
    \item H2: Fine-tuning with the target language will significantly improve MLLM performance. That is, fine-tuning a model on Spanglish will maximize its performance in identifying ideas expressed in Spanglish. 
\end{itemize}
During translanguaging, speakers draw from their entire linguistic background and cultural identity to navigate and defy the socio-political boundaries of "proper" language use, in turn promoting authentic social interaction and communication~\cite{hamman2018translanguaging}. Additionally, Li Wei~\cite{WEI20111222} defines the term \textit{translanguaging space} as a place where cultural and linguistic boundaries eclipse in bilinguals' daily lives. Within these spaces, both criticality and creativity work simultaneously to inform bilingual learners' perceptions on socio-cultural and linguistic phenomena and determine what constitutes the norms of language use.

Much of the discourse surrounding these practices has shifted towards how to leverage students’ linguistic resources in learning environments. With increasing globalization, instructional resources such as the \textit{CUNY-NYSIEB Translanguaging Guides} now include pedagogical strategies that use bilingualism as a resource to leverage learning. These guides emphasize collaborative work and provide linguistic resources for EBLs of varying age and grade levels, and have been widely used in the past decade by educational communities working towards putting theory into practice~\cite{hesson2014translanguaging}. These resources have useful applications when promoting students’ existing language background, cultural identity, and multi-modal practices (i.e., gesture) as meaning- and sense-making processes. As the learning analytics field consistently showcases diverse learner perspectives, we view translanguaging as an increasingly valuable framework that empowers bilingual learners to convey ideas and understanding using their complete linguistic repertoire.

\subsection{LLM Support in Education}

Large Language Models (LLMs), with their ability to generate language from large-scale datasets, have gained prominence in the learning analytics field. They are applied in tasks such as essay scoring and feedback generation, highlighting LLMs' growing role in saving instructor time and providing valuable feedback for learners~\cite{li2024applying}. For instance, Automated Essay Scoring (AES) systems reduce examiner workloads in large-scale assessments like TOEFL and GMAT. LLMs like BERT outperformed previous AES methods, establishing new benchmarks in the field~\cite{hirao-etal-2020-automated}. Beyond scoring, LLMs are capable of providing personalized feedback for intelligent tutoring systems~\cite{MEYER2024100199}, aiding researchers in developing high-quality educational tools. They also show aptitude in classroom-specific subjects such as mathematics, are capable of generating multiple-choice questions~\cite{mcnichols2023automated}, plotting figures~\cite{bulusuautomated}, and provide teacher training via simulations~\cite{LeeGenerativeAF}. With these advancements, it is evident that LLMs have immense potential in classrooms, making it increasingly important to extend the same support to bi/multilingual classrooms.

\subsection{Multilingual LLMs}
In general, since LLMs are trained on English-centric data, LLMs perform better in English than in non-English languages~\cite{zhu2023multilingual}. However, newer MLLMs such as GPT-4~\cite{achiam2023gpt} demonstrate increasing multilingual capabilities and are outperforming their predecessors on multilingual benchmarks~\cite{ahuja2023megaverse}. Despite recent advancements in MLLM research, further inspection reveals a challenge: many MLLMs still struggle with code-switching due to the lack of diverse linguistic resources in training data~\cite{zhang2023multilingual}. Additionally, there is a lack of comprehensive benchmarks to support and analyze MLLMs recent developments~\cite{qin2024multilingual}. As a result, many languages with limited digital resources are excluded~\cite{joshi2020state}. Code-switching resources in particular are scarce due to the lack of large-scale annotations that require multilingual human-raters~\cite{winata2022decades}. To address this data shortage, researchers have used manually created datasets using methods such as random replacements and noun-phrase translations~\cite{laureano2024code} or by using MLLMs to generate synthetic code-switched texts~\cite{hu2023improving, yong2023prompting}. In synthetic generation, recent studies show that newer MLLMs such as GPT-4 are more robust for code-switching and cross-lingual understanding~\cite{huzaifah2024evaluating}, posing a potential solution for contexts requiring code-switching capabilities. As bilingual learners engage in classroom discussions in two or more languages, it is crucial to develop fair and reliable datasets for MLLMs to capture dynamic bi/multilingual processes.

\section{Methods}
\label{sec:methodology}

To test our hypotheses of MLLM performance across monolingual (Spanish, English) and bilingual (Spanglish) settings, we developed parallel datasets across English, Spanish and Spanglish. We prepared our primary dataset in English and then translated it into Spanish and Spanglish. Pre-trained LLMs are models already trained on large amounts of data and can be fine-tuned for specific tasks. To test whether pre-trained models exhibited worse performance in bilingual languages, as H1 suggests, we evaluated the grading performance of two MLLMs across all three languages. For H2, which hypothesizes target-language fine-tuning as the most effective way to improve performance for a language, we performed two experiments: one aimed at improving Spanglish performance through fine-tuning, and another to explore cross-lingual transfer. We first outline our synthetic data preparation process, then describe our experimental designs.

\subsection{Synthetic Data Preparation} 

We used Claude 3.5 Sonnet \cite{anthropic2024claude35}, a large proprietary (closed-source) language model to generate our synthetic datasets, which we used for training and evaluation. 

First, we prompted the model to generate question-answer pairs in English along with a binary grade of \textit{Acceptable} or \textit{Unacceptable}. We focused on Science and Social Science topics for grades 6 through 10. Without careful prompting, Claude tends to generate low-quality data, producing \textit{Unacceptable} answers that are both unrealistic and significantly shorter than \textit{Acceptable} ones. The questions also clustered around the same limited topics, allowing the MLLM grader to use shortcuts and falsely inflate performance. To tackle this, we explicitly prompted the model to generate realistic and nuanced unacceptable answers, maintain similar average lengths for acceptable and unacceptable answers, and cover a wide range of topics. We generated 10 question-answers at a time since Claude tends to produce low-quality examples when asked to output in bulk. This approach also allowed us to continuously monitor the quality of data being produced and re-prompt if necessary. Finally, we conducted a human evaluation to verify the accuracy of acceptable and unacceptable classifications. Two annotators reviewed a randomly sampled subset of 100 data points, representing 10\% of the entire dataset. Using binary labels, the annotators indicated if they agreed with the model's labels. To help validate the use of our synthetic labels, the annotators rated correctness as 99\% and 98\%, with the percentage overlap between them being 97\%.

We used Claude again to translate the answer portion of the dataset into Spanish and Spanglish. For Spanglish, we periodically prompted the model to vary the English:Spanglish ratio to ensure diverse code-switching examples. To assess the Spanglish quality, we conducted human evaluation to verify translation Accurateness (0-inaccurate translation, 1-accurate translation) and Naturalness (0-unnatural, 1-natural: annotators may see themselves or another bilingual person speaking it), similar to the approach in \cite{yong2023prompting}. Two bilingual annotators, fluent in English and Spanish, received a dataset of 100 samples, with 50 of them overlapping. Individually, they rated 96\% and 100\% for Accurateness, and both rated 100\% for Naturalness. Overall, across the 150 unique data points annotated, we achieved an average score of 96.7\% Accurateness and 98.7\% Naturalness. For 50 overlapping samples, our two annotators showed high agreement, with a percentage overlap of 96\% for Accurateness and 100\% for Naturalness.

We structured our dataset with a train-validation-test split of 150-150-1000. Our test set contained an equal distribution of Science and Social Science topics across grades 6 to 10. While prior research has used synthetic data to improve performance for text classification \cite{li2023synthetic} and generate code-switched data \cite{yong2023prompting, laureano2024code}, our study to the best of our knowledge is the first to use synthetic code-switched data to improve MLLM grading performance on bilingual student writing.

\subsection{Experimental Design}

We performed fine-tuning and evaluation using open-source language models to avoid sending the data to a third-party server -- a critical factor given the importance of privacy in educational data. Smaller open-source models like Llama 3.1 are free to use and fine-tune, requiring significantly lower computational resources. Specifically, we used the ‘instruct’ versions of Llama 3.1 (8 billion parameters) \cite{meta2024_llama31} and Mistral NeMo (12 billion parameters) \cite{mistral2024nemo}. A consistent grading prompt was used across both models, all languages and all experimental setups. The prompt detailed the task: grade single-sentence answers as acceptable or unacceptable.

\subsubsection{Zero-shot Baseline for English, Spanish and Spanglish}

To test \textbf{H1}, we compared the grading performance of Llama 3.1 and NeMo across English, Spanish and Spanglish in a zero-shot setting. Zero-shot learning refers to the model's ability to perform a task without having seen any training examples for it. This approach helped identify potential language biases in the two models and served as a baseline for our next experiment.

\subsubsection{Improving Spanglish Performance}

To evaluate whether fine-tuning with the target language improves Spanglish performance, we compared the zero-shot, few-shot prompting, and fine-tuned performance of Llama 3.1 and NeMo for Spanglish. Few-shot learning is when we provide the model with a few examples of our task as part of the prompt. For few-shot, we provided three Spanglish question-answer-grade examples and evaluated performance using the pre-trained models. Fine-tuning involves adapting a pre-trained model for a particular task by training it with a task-specific dataset. For fine-tuning, we used Unsloth's open-source library~\cite{unslothai_unsloth_2023} to accelerate training with LoRA (Low-Rank Adaptation) \cite{hu2021loralowrankadaptationlarge} and reduce memory usage. We fine-tuned both models with the synthetic Spanglish training set (150 samples) for 3 epochs with a learning rate of 2e-4.

\subsubsection{Cross-lingual Transfer}


\begin{figure}
    \centering
    \includegraphics[width=1\linewidth]{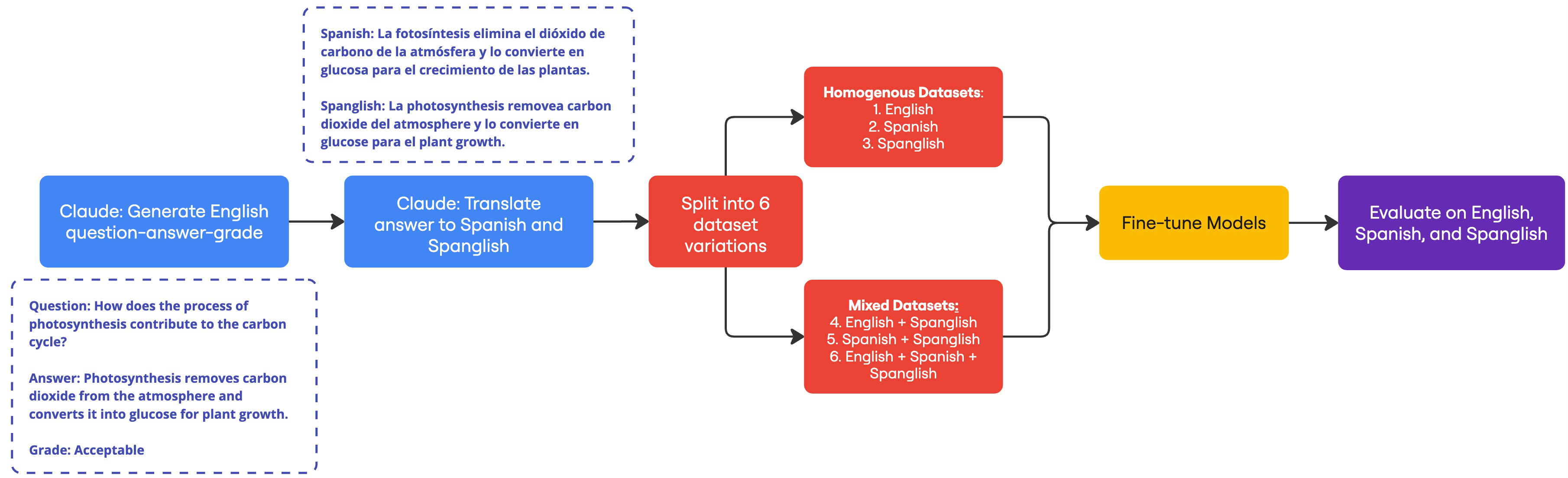}  
        \caption{Experimental Pipeline for evaluating cross-lingual transference}
    \label{fig:cross-lingual-eval}
\end{figure}

In \textbf{H2}, we hypothesized that fine-tuning with the target language is the best way to improve performance for that language. Specifically, we expected Spanglish fine-tuning to be the most effective way to improve Spanglish performance compared to English or Spanish fine-tuning. To investigate how fine-tuning in each language influences performance in other languages (cross-lingual transfer), we fine-tuned Llama 3.1 on each language and evaluated its performance across all three languages (see Figure \ref{fig:cross-lingual-eval}). To improve and balance the performance across the three languages, we also fine-tuned Llama 3.1 on three mixed datasets: English (100) + Spanglish (50), Spanish (100) + Spanglish (50), English (50) + Spanish (50) + Spanglish (50). The mixed language datasets were prepared by sampling and merging data from the three monolingual training datasets, maintaining the same dataset size.  All the six fine-tuned models were tested on English, Spanish and Spanglish.  Additionally, we computed and compared the average performance of each model across languages.

\section{Results}
\label{sec:results}


Our analysis investigated potential bias in MLLMs by comparing the assessment of bilingual and monolingual writing. We also focused on methods to enhance bilingual performance and examine cross-lingual transfer after fine-tuning. 

Here we present the results of our three experimental setups: \textit{3.1 Zero-shot Baseline}, \textit{3.2 Improving Spanglish performance}, and \textit{3.3 Cross-lingual Transfer}. In \textit{3.1}, we investigated our first hypothesis, \textbf{H1}, which posits there is bias in the performance of MLLMs when student responses are in Spanglish compared to English and Spanish. \textit{3.2} and \textit{3.3} explored our second hypothesis, \textbf{H2}, which seeks to understand the differences in MLLMs performance on English, Spanish and Spanglish when fine-tuned using various languages for the same task. For all experiments, we used Area Under the ROC Curve (AUC) as our performance metric.

\subsection{Zero-shot Baseline} 
       

     

\begin{figure}[htbp]
  \centering
  \vskip 0.1in
  \begin{minipage}[tb]{0.48\textwidth}
    \centering
    \includegraphics[width=0.95\textwidth]{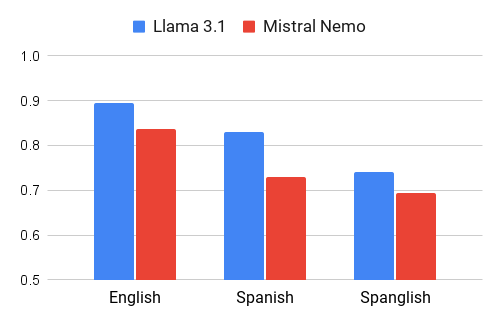}
    \vskip 0.1in
    \caption{Comparing pre-trained performance (AUC) of Llama 3.1 and Mistral NeMo for English, Spanish, and Spanglish grading. Both models perform best for English and worst for Spanglish
    }
    \label{fig:baseline_eval}
    \vskip -0.1in
  \end{minipage}
  \hfill
  \begin{minipage}[tb]{0.48\textwidth}
    \centering
    \includegraphics[width=0.95\textwidth]{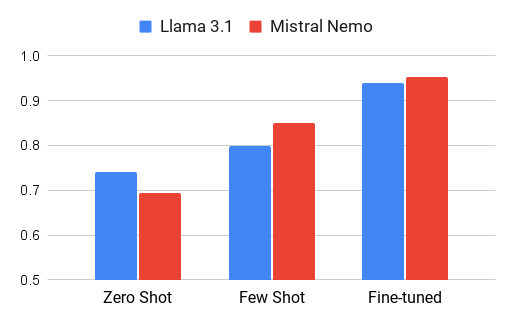}
    \caption{Comparing Spanglish grading performance (AUC) for Llama 3.1 and Mistral NeMo using zero-shot prompting, few-shot prompting, and fine-tuning using Spanglish. There is significant performance gains from zero-shot to few-shot to fine-tuning for both models.  }
    \label{fig:baseline_methods}
  \end{minipage}
\end{figure}

Figure~\ref{fig:baseline_eval} records the grading performance of pre-trained models with zero-shot prompting in three languages. We observed the same performance hierarchy in both models, with English exhibiting superior results, followed by Spanish, while Spanglish yielded the lowest scores. We also see for all of the languages, Llama 3.1 performed better than NeMo despite being the smaller model. Across the three languages, Llama 3.1 and NeMo achieved average AUC scores of 0.82 and 0.75 respectively. The breakdown by language shows performance in English (0.90 vs 0.84), Spanish (0.83 vs 0.73), and Spanglish (0.74 vs 0.69). The results strongly support our first hypothesis, \textbf{H1}, that there is a bias in MLLM performance across bilingual and monolingual languages, even for high-resource language pairs. This result recommends not to use these models in their pre-trained state for bilingual assessment.



\subsection{Improving Spanglish Performance} Figure~\ref{fig:baseline_methods} illustrates the performance of Llama 3.1 and NeMo on Spanglish using zero-shot prompting, few-shot prompting, and fine-tuning approaches. As expected, both models exhibited superior performance after fine-tuning compared to few-shot and zero-shot prompting. We used three graded Spanglish Q\&A examples for few-shot learning in conjunction with the zero-shot prompt, while fine-tuning was done on 150 graded Spanglish Q\&A samples. 

Although NeMo under-performed compared to Llama 3.1 for baseline tests, it received a significant performance enhancement with few-shot learning. Specifically, NeMo showed a substantial leap from 0.69 in zero-shot to 0.85 with few-shot prompting, while the performance of Llama 3.1 increased from 0.74 to 0.8. After fine-tuning, both Llama 3.1 (0.94) and NeMo (0.95) performed similarly. This represents a considerable improvement over zero-shot learning (27\% for Llama, 37.7\% for NeMo) and few-shot learning (17.5\% for Llama, 11.8\% for NeMo), highlighting the effectiveness of fine-tuning with the target language and supporting \textbf{H2}.



\subsection{Cross-lingual Transfer} 

In this experiment, we analyzed the performance of Llama 3.1 after fine-tuning on six datasets: three homogeneous (English, Spanish and Spanglish) and three mixed combinations of these languages (English + Spanglish, Spanish + Spanglish, and English + Spanish + Spanglish). For fair comparison, the six datasets included the same Q\&A examples. We tested each model for English, Spanish and Spanglish. In this section, we compare the average performance of the models for all three languages and conduct a fine-grained analysis on language-wise performance.

\begin{table}[h]
  \caption{Comparing average performance across English, Spanish and Spanglish for the six fine-tuned configurations. Spanglish fine-tuned and English-Spanglish fine-tuned give best performance on average.}
  \label{tab:avg-performance}
  \begin{tabular}{lc}
    \toprule
    \textbf{Llama 3.1 Model} & \textbf{Average AUC} \\
    \midrule
    Baseline & 0.822 \\
    English Fine-tuned & 0.885 \\   
    Spanish Fine-tuned & 0.910 \\   
    Spanglish Fine-tuned & 0.955 \\
    English-Spanglish Fine-tuned & 0.955 \\
    Spanish-Spanglish Fine-tuned & 0.948 \\
    English-Spanish-Spanglish Fine-tuned & 0.945 \\
    \bottomrule
\end{tabular}
\end{table}

\begin{figure}[ht]
    \centering
    \includegraphics[width=0.45\textwidth]{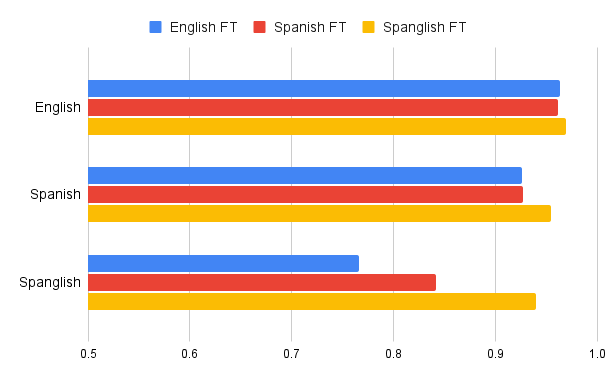}   \includegraphics[width=0.45\linewidth]{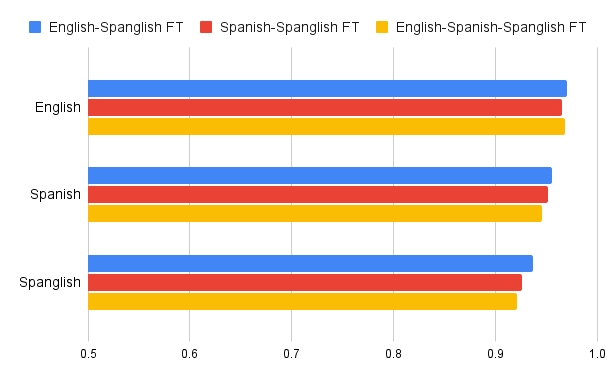}
    \caption{Comparing performance (AUC) of fine-tuning Llama 3.1 in different languages (left) and language combinations (right) evaluated on English, Spanish and Spanglish. Spanglish demonstrates robust cross lingual transfer, while English and Spanish fine-tuned models struggle with Spanglish. Combination models perform similarly to each other and perform well for all languages.}
    \label{fig:cross_ling_fts}
\end{figure}


Table~\ref{tab:avg-performance} shows the average performance of the six fine-tuned models and pre-trained Llama 3.1 (baseline) for the three languages. We saw that all fine-tuned models performed significantly above the baseline (> 0.822), with Spanglish and the combination models performing particularly well (> 0.945). Spanglish-only fine-tuned and English-Spanglish fine-tuned models outperformed every other fine-tuned Llama 3.1 model, both averaging 0.955. They are closely followed by the other mixed models (0.948 for Spanish-Spanglish and 0.945 for English-Spanish-Spanglish) and then by Spanish-only (0.910) and English-only (0.885) models. However, comparing the average performance does not give the full picture as most of the difference arises from performance in Spanglish, which we will see in the following part of this analysis. 

Figure~\ref{fig:cross_ling_fts} depicts the performance of the six models for each language. Among homogeneous models, the Spanglish fine-tuned model exhibited the most robust cross-lingual transfer. It not only delivered the optimal performance for Spanglish (0.940) among the six models, but also achieved near-optimal scores for English (0.970) and Spanish (0.955). English fine-tuned and Spanish fine-tuned models significantly improved in performance from their respective baselines (0.895 to 0.964 for English, 0.830 to 0.927 for Spanish). They also displayed strong transferability amongst themselves, with the English model improving Spanish performance compared to the Spanish model (0.926 and 0.927, respectively), and vice versa (Spanish model on English: 0.962; English model on English: 0.964). However, both models under-performed on Spanglish with AUC scores of 0.766 for the English model and 0.842 for the Spanish model. The performance of the English model on Spanglish was just around the Spanglish baseline (0.740), but the Spanish model improved Spanglish performance by roughly 13\%. Overall, we found that while target-language fine-tuning improved MLLM performance for all languages, it may not necessarily yield the best performance, contrary to H2. The Spanglish fine-tuned model performed better on English and Spanish than the target-language fine-tuned models. We also saw there was strong cross-lingual transfer between Spanglish to English and Spanish, as well as bi-directionally between English and Spanish. However, fine-tuning with English gave only minimal improvements for Spanglish performance, and the Spanish model improved the performance by only a moderate amount. 

On average, mixed models performed better than the homogeneous models. We believe this difference was due to the mixed models including Spanglish samples in fine-tuning. Even though the mixed datasets had only 50 Spanglish samples, their Spanglish performance was significantly better than English-only and Spanish-only models and close to the Spanglish-only model. Specifically, Spanglish performance for English-Spanglish (0.937), Spanish-Spanglish (0.926), and English-Spanish-Spanglish (0.921) models are comparable to the optimal Spanglish fine-tuned model (0.940). For English performance, all three mixed models slightly outperformed the English-only fine-tuned model (0.964) with the English-Spanglish model achieving the optimal performance (0.971). Similarly, for Spanish, the mixed fine-tuned models: English-Spanglish (0.956), Spanish-Spanglish (0.952), and English-Spanish-Spanglish (0.946), exceeded the Spanish-only model (0.927). In summary, mixed combination models performed better for English and Spanish than their respective fine-tuned models, and significantly better than English-only and Spanish-only models for Spanglish.

Overall, Spanglish fine-tuned and English-Spanglish fine-tuned models gave the best results. Fine-tuning, especially with the target language, is an effective way to improve performance for the language; however, it did not necessarily give the best performance. For English and Spanish, the Spanglish fine-tuned model and mixed combination models outperformed target language fine-tuning. Hence, the results of experiment 3.3 partially support \textbf{H2}, with fine-tuning the target language being a robust way to improve performance for that language.

\section{Discussion}
\label{sec:discussion}

In this study, we looked at the performance of open-source MLLMs in assessing bilingual student writing. We investigate Llama 3.1 (8B) and Mistral NeMo (12B) under zero-shot, few-shot and fine-tuned conditions to identify differences in Spanglish grading performance compared to English and Spanish. We confirm our first hypothesis that there is a bias in grading performance for bilingual writing (Spanglish) compared to monolingual writing (English and Spanish). Our findings partially support our second hypothesis: fine-tuning with the target language significantly enhances grading performance, but does not necessarily give the best performance.

In zero-shot conditions, both models perform as expected, with English grading achieving the best results, followed by Spanish then Spanglish, with considerable performance gaps between all three. Low accuracy in Spanglish assessment indicates that current smaller open-source MLLMs are unreliable for bilingual assessment. Spanish performance also lags significantly behind English for pre-trained models, indicating a need to make MLLMs more inclusive to non-English languages.

However, fine-tuning with Spanglish substantially enhances performance on Spanglish assessment, outperforming few-shot and zero-shot approaches. It also exhibits robust generalization to both Spanish and English data, achieving near-optimal scores for their grading performance. Although fine-tuning with Spanish and English demonstrates effective cross-lingual transfer between the two languages, both models under-perform for Spanglish grading. In fact, the English fine-tuned model performs approximately at the Spanglish baseline. We believe this is caused by a homogeneity in training data: with minimal exposure to hybridized languages (Spanglish), even fine-tuned models can fail at processing bilingual data.

Mixed-dataset fine-tuning is more consistent across languages compared to homogeneous fine-tuning. This is likely due to the inclusion of Spanglish in all mixed-data combinations (English + Spanglish, Spanish + Spanglish, English + Spanish + Spanglish) to address the performance gap seen in monolingual fine-tuning. Overall, their performance is comparable to or very close to that of the Spanglish-only fine-tuned model. Notably, mixed datasets achieved significant results with only 50 Spanglish samples (versus 150 in the pure Spanglish dataset), suggesting substantial grading performance gains are possible with limited bilingual data.

Importantly, our findings have implications for learning analytics research. First, they underscore the need for a more robust measurement, collection, and analytic process of mixed-language and code-switching data using authentic bilingual corpora. Second, as current open-source MLLMs show limitations in assessing bilingual inputs, our results encourage the use of mixed-language datasets to enhance the reliability of MLLM grading. By doing so, learning analytics approaches that utilize MLLMs can better adhere to bilingual learners’ diverse linguistic contexts and translanguaging practices, ultimately improving both the quality of assessments and better reflecting the environments where learning takes place. 

As research on MLLMs in educational settings become more prevalent, it is essential that their uses are inclusive and supportive of bilingual learners. Smaller, open-source models are often preferable for education use cases due to their cost-effectiveness, lower computing requirements, ease of customization, and better data privacy, ownership, and governance. However, these models under-perform for bilingual tasks, even for high-resource language pairs like Spanglish. It is promising that they demonstrate significant improvements upon fine-tuning with bilingual data and exhibit strong cross-lingual transferability between languages, although these patterns need to be confirmed with authentic real-world data. The main limitation of our analysis is the reliance on synthetic data for evaluation, mainly due to the lack of publicly available datasets for bilingual assessment. Our study calls for further research to explore the potential of MLLMs in bilingual education, particularly for more complex assessment tasks, and to investigate effectiveness of MLLMs grading performance with additional low-resource languages.

In conclusion, we position fine-tuning as an effective strategy for improving MLLMs grading performance on mixed-language datasets. Attending to the natural process of translanguaging (i.e., Spanglish) helps us adapt MLLMs to be more suitable to the diverse linguistic backgrounds of bilingual students. We present both the average performance and fine-grained analysis on different languages and language combinations as templates for other researchers. These findings elucidate the performance enhancement and cross-lingual transferability obtained by fine-tuning with bilingual data. As learning analytics provides technologies to support diverse learners, continual investigation of potential biases in MLLMs is vital.

\begin{acks}
\rucomment{Acknowledgements added here after anonymity is removed.}
\end{acks}

\bibliographystyle{ACM-Reference-Format}
\bibliography{references}

\appendix


\end{document}